\documentclass[letterpaper, 10 pt, conference]{ieeeconf}  
\IEEEoverridecommandlockouts                              
\usepackage[noadjust]{cite}

\usepackage{amsmath}
\usepackage{yhmath}
\usepackage{mathtools} 
\usepackage{amssymb} 
\usepackage{sidecap}
\usepackage{graphicx}
\usepackage{subcaption}
\usepackage{multirow}
\usepackage{epsfig}
\usepackage{psfrag}
\usepackage{caption}
\usepackage[super]{nth} 
\usepackage[normalem]{ulem} 
\usepackage{color}
\usepackage{xcolor} 
\usepackage{booktabs} 
\usepackage{siunitx} 
\usepackage{booktabs} 
\usepackage{colortbl} 
\usepackage{textcomp} 
\usepackage{tikz}
\usepackage[noadjust]{cite} 

\usepackage{wasysym}  

\newcommand{\RNum}[1]{\uppercase\expandafter{\romannumeral #1\relax}} 

\newcommand{\realfield}[1]{\hbox{I \kern -.5em R}^{#1}}
\newcommand {\mb}[1]{\mathbf{#1}}
\newcommand {\bs}[1]{\boldsymbol{#1}}
\newcommand{\uvec}[1]{\hat{\mathbf{#1}}}
\newcommand{\Rot}[2]{{^{#1}\mathbf{R}}_{#2}}  
\newcommand{\T}{^{\mathrm{T}}}  

\definecolor{LightGray}{gray}{0.9}
\newcolumntype{a}{>{\columncolor{LightGray}}l}
\usepackage{balance}

\usepackage{color}  

\usepackage[textsize=scriptsize, bordercolor=white,backgroundcolor=gray!30,linecolor=black,colorinlistoftodos]{todonotes}  
\usepackage[normalem]{ulem} 
\usepackage{soul} 

\title{\LARGE \bf
Kinematic Modeling and Compliance Modulation of \\ Redundant Manipulators Under Bracing Constraints
}
\author{Garrison~L.H.~Johnston$^{1}$,~Andrew~L.~Orekhov$^{1}$,~Nabil~Simaan$^{1}$$^{\dag}$ 
\thanks{$\dag$ Corresponding author}
\thanks{$^{1}$Department of Mechanical Engineering, Vanderbilt University, Nashville, TN 37235, USA
        {\tt ( garrison.l.johnston, andrew.orekhov, nabil.simaan) @vanderbilt.edu}}
\thanks{This work was supported by NSF award \#1734461 and by Vanderbilt internal university funds. A. Orekhov was partially supported by the NSF Graduate Research Fellowship under \#DGE-1445197.}}

\makeatletter
\let\NAT@parse\undefined
\makeatother
\usepackage{hyperref}  
\usepackage{fancyhdr} 

\usepackage{cleveref}
\begin{document}
\maketitle
\thispagestyle{empty}

\thispagestyle{fancy}
\fancyhf{}
\renewcommand{\headrulewidth}{0pt}
\lhead{2020 IEEE International Conference on Robotics and Automation (ICRA). Accepted Version. }
\rfoot{\centering \scriptsize \copyright 2020 IEEE. Personal use of this material is permitted. Permission from IEEE must be obtained for all other uses, in any current or future media, including reprinting/republishing this material for advertising or promotional purposes, creating new collective works, for resale or redistribution to servers or lists, or reuse of any copyrighted component of this work in other works.}

\pagestyle{empty}
\begin{abstract}
Collaborative robots should ideally use low torque actuators for passive safety reasons. However, some applications require these collaborative robots to reach deep into confined spaces while assisting a human operator in physically demanding tasks. In this paper, we consider the use of in-situ collaborative robots (ISCRs) that balance the conflicting demands of passive safety dictating low torque actuation and the need to reach into deep confined spaces. We consider the judicious use of bracing as a possible solution to these conflicting demands and present a modeling framework that takes into account the constrained kinematics and the effect of bracing on the end-effector compliance. We then define a redundancy resolution framework that minimizes the directional compliance of the end-effector while maximizing end-effector dexterity. Kinematic simulation results show that the redundancy resolution strategy successfully decreases compliance and improves kinematic conditioning while satisfying the constraints imposed by the bracing task. Applications of this modeling framework can support future research on the choice of bracing locations and support the formation of an admittance control framework for collaborative control of ISCRs under bracing constraints. Such robots can benefit workers in the future by reducing the physiological burdens that contribute to musculoskeletal injury.
\end{abstract}


\section{Introduction} \label{ch:intro}
 \par  The deployment of collaborative robots in confined spaces (such as in service and repair of airplane wings and household crawl spaces) can substantially benefit workers in terms of reduction of physiological burden and the associated risks of musculoskeletal disorders \cite{Lorenzini2019}. Physical human-robot interaction with such robots can allow the workers to remain in control of critical aspects of service and repair tasks while reducing their physiological burden. We call these robots \textit{in-situ collaborative robots} (ISCRs) and we believe they will allow rapid deployment and use within semi-structured environments while avoiding the potential pitfalls of the following two alternatives. The first alternative, complete automation, requires exact knowledge of the environment and comes at a cost of increased burden in environment mapping, robot registration to the environment, task programming, and limited repertoire of tools suitable for rapid tool exchange. Telemanipulation of service robots is the second alternative which places the worker outside the confined space, but comes at a cost of limited sensory presence, limited situational awareness, and increased cost of the robotic setup.
\par The collaborative use of ISCRs in confined spaces requires both active (e.g. safe collision detection and avoidance) and passive measures of safety (e.g. safety in case of collision). To support passive safety, these robots must avoid the use of large torque actuators, minimize link inertia, and limit acceleration. This requirement for minimal torque actuation, however, comes in stark contrast to many application scenarios requiring service and repair in deep confined spaces such as airplane wings. Such tasks require long-reach robots, which need high torque actuators to support their self-weight. This opposes passive safety requirements. To overcome this challenge and increase end-effector stiffness, a combination of static balancing and bracing may be used. For example, \cite{Diken1995} demonstrated the potential benefits of static balancing for reducing torque requirements on the PUMA robot. Also, \cite{Fang2019} demonstrated the benefits of bracing for actuator torque reduction. This paper will focus on bracing as a solution to the aforementioned design tradeoffs.
\par Relevant works on bracing include early works of Book \cite{BookBrcing1984} who first proposed bracing and considered attachments for bracing. In \cite{HollisBracingAssembly1992}, Hollis presented the concept of macro-micro manipulation with the macro manipulator using bracing for supporting accurate assembly. Robot dynamics with bracing constraints and link flexibility was also considered in \cite{Book_bracing_kin1994}. Delson and West \cite{Delson1993} modeled the effect of bracing on the natural frequency of serial manipulators. Lee and Kim considered reconfigurable systems with bracing constraints and formulated their force and kinematic dexterity ellipsoids \cite{Lee1991bracing_ellipsoids}. They also formulated their dynamic manipulability in \cite{KimDynamicManipBracing1991}. West and Asada \cite{WestAsadaBracingForceControl1985} used virtual closed-loop linkages to represent the constrained kinematics with a single contact and assumed perfect knowledge of the task and contact of a rigid robot and environment. This formulation was subsequently generalized with a screw-theoretic approach by Featherstone et al. \cite{FeatherstoneFrictionlessContactHybridControl1999} for multiple frictionless contact constraints. Multi-contact control problems have been considered in \cite{ParkMultiContactForceControl2006,Wang2010a,Wang2010,Itoshima2011,Washino2012}.
\par Relative to prior work, this paper aims to present a kinematic and compliance modeling framework suitable for kinematic redundancy resolution for modulating end-effector compliance under bracing constraints. To our knowledge, a compliance modulation redundancy resolution strategy that accounts for different kinematic bracing constraints and a modeling of the effect of bracing on task-specific end-effector compliance has not been presented. Previous works on stiffness modulation considered actuation redundancy in parallel robots \cite{yi_geometric_1993, simaan_geometric_2003, jamshidifar_kinematically-constrained_2017, kock_parallel_1998, chakarov_study_2004, muller_stiffness_2006}, joint-level compliance control in serial robots \cite{kim_analysis_1997, lee_upper-body_2014, lim_design_2013}, use of kinematic redundancy for modulating stiffness \cite{simaan_stiffness_2003, abdolshah_optimizing_2017, anson_orientation_2017}, and a combination of joint-level stiffness control and kinematic redundancy \cite{alamdari_stiffness_2018,rice2018passive,orekhov_directional_2019}.
\par The contribution of this work, given in Sections \ref{ch:kinematics} and \ref{ch:stiffness}, is a kinematic and compliance modeling framework for a serial robot under bracing constraints to allow the exploration of potential tradeoffs of bracing at a particular location along a robot. This modeling framework enables a redundancy resolution strategy, which we provide in Section \ref{ch:red_res}, that respects bracing constraints while modulating end-effector compliance to improve the execution of a given task. Our kinematic simulations in Section \ref{ch:results} demonstrate that bracing can reduce compliance and that the redundancy resolution strategy we present can even further reduce compliance while satisfying the kinematic bracing constraints and providing improved kinematic conditioning.

\section{Modeling nomenclature}
In the following paper we will consistently use the following notation:
\begin{itemize}
\item $\{0\}$: world frame with its origin arbitrarily chosen at the base of the robot.
\item $\{a\}$: a frame having its origin at point $\mb{a}$ and its orientation given by rotation matrix $^{0}\mb{R}_a$.
\item ${}^b\hat{\mb{x}}_a$, ${}^b\hat{\mb{y}}_a$, ${}^b\hat{\mb{z}}_a$: unit vectors of frame $\{a\}$ expressed in $\{b\}$.
\item $^{b}\mb{R}_a$: orientation of $\{a\}$ relative to $\{b\}$.
\item $\Delta^{b}\mb{x}_a$: a twist comprised of the linear velocity followed by angular velocity and expressed in a frame centered at $\mb{a}$ and parallel to $\{b\}$.
\item  $^{b}\mb{w}_a$: a wrench comprised of the force followed by the moment and expressed in a frame centered at $\mb{a}$ and parallel to $\{b\}$.
\item $\mb{J}$: geometric instantaneous direct kinematics Jacobian. Recall that this Jacobian relates joint speeds to end-effector twist defined at a frame parallel to the robot's base frame, but centered at the end-effector point.
\item $\left[\cdot\right]_\times$:  the skew symmetric cross product operator. For example, given a vector $\mb{a} = \begin{bmatrix} a_1 & a_2 & a_3 \end{bmatrix}\T$, the skew symmetric cross product matrix of $\mb{a}$ is:
\begin{equation}\label{eq:skew_sym}
  \left[\mb{a}\right]_\times = \begin{bmatrix}
                          0   & -a_3 & a_2 \\
                          a_3 & 0    & -a_1 \\
                         -a_2 & a_1  & 0
                        \end{bmatrix}
\end{equation}
\end{itemize}

\section{Constrained Kinematic Modeling}\label{ch:kinematics}
\begin{figure}[htbp]
    \centering
    \includegraphics[width=0.8\columnwidth]{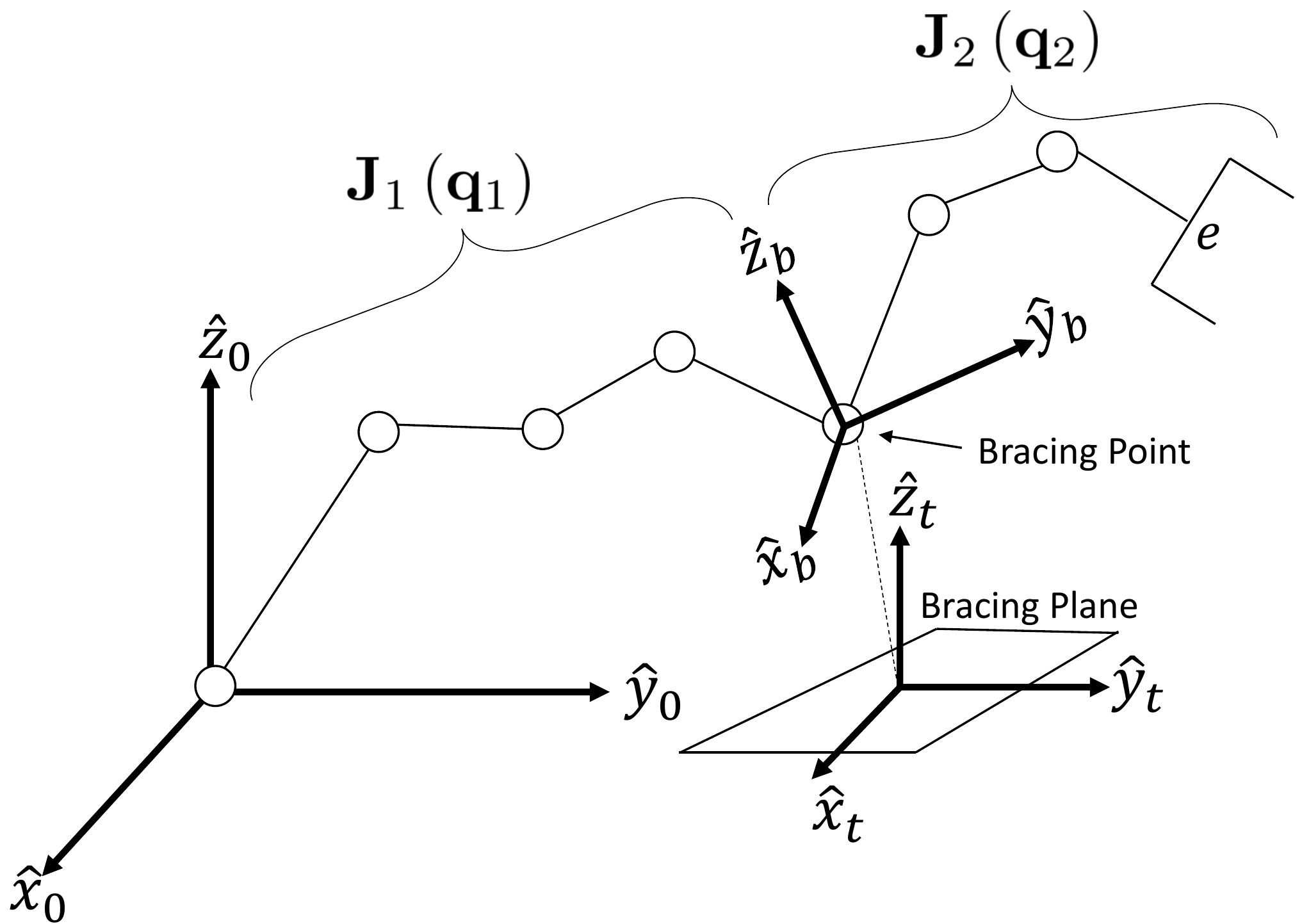}
    \caption{Example serial robot with frame $\{b\}$ located at the bracing point and frame $\{t\}$ located on the bracing plane. In this figure, the bracing plane has been drawn separated from the bracing point for visual clarity.}
    \label{fig:example_robot}
\end{figure}
In this section, we present a kinematic model for serial robots with a single bracing contact along their body at point $\mb{b}$. As shown in Fig. \ref{fig:example_robot}, a serial manipulator bracing against the environment can be broken into two kinematic chains. The first kinematic chain spans from the origin of the world frame $\{0\}$ to the origin of frame $\{b\}$. The second kinematic chain spans from $\mb{b}$ to the end-effector $\mb{e}$. In free space, the kinematic chains have the following instantaneous kinematics:
\begin{subequations}
\begin{equation}\label{eq:J1}
  \Delta{}^0\mb{x}_b = \mb{J}_1\dot{\mb{q}}_1
\end{equation}
\begin{equation}\label{eq:J2}
 {}\Delta^b\mb{x}_{e/b} = \mb{J}_2\dot{\mb{q}}_2
\end{equation}
\end{subequations}
\noindent where $\mb{J}_1$ and $\mb{J}_2$ are the geometric Jacobians of the first and second kinematic chains, respectively. $\Delta{}^0\mb{x}_b$ is the twist of the bracing point expressed in a frame parallel to $\{0\}$ and located at $\mb{b}$. ${}\Delta^b\mb{x}_{e/b}$ is the twist of the end-effector relative to $\{b\}$ and expressed in a frame having its origin at  $\mb{e}$ and parallel to $\{b\}$.

\par When the robot is braced, the bracing point is constrained to move in a set of allowable velocities $M\subset\realfield{l}$. This constraint can be represented with a matrix $\mb{H}\in\realfield{6\times l}$ whose columns are unit twists which form a basis for $M$. The instantaneous velocity of $\mb{b}$ can be truncated as a vector $\dot{\widetilde{\mb{b}}}\in\realfield{l\times1}$ whose elements are velocities in the allowable directions, i.e. in terms of the basis formed by the columns of $\mb{H}$ \cite{Murray1994}. Using these matrices, the bracing constraint can be represented as:
\begin{equation}\label{eq:constraint}
  \Delta{}^0\mb{x}_b = \mb{H}\dot{\widetilde{\mb{b}}}
\end{equation}
For a frictionless point contact, the bracing contact point is free to rotate about all directions and translate in the ${}^0\hat{\mb{x}}_t$ and ${}^0\hat{\mb{y}}_t$ directions where frame $\{t\}$ is a predefined frame tangent to the constraint surface at the current location of the bracing point. The $\mb{H}$ matrix in this case is:
\begin{equation}\label{eq:H_point}
  \mb{H} = \begin{bmatrix}
                 {}^0\hat{\mb{x}}_t & {}^0\hat{\mb{y}}_t & \mb{0}             & \mb{0}             & \mb{0}\\
                 \mb{0}             & \mb{0}             & {}^0\hat{\mb{x}}_t & {}^0\hat{\mb{y}}_t & {}^0\hat{\mb{z}}_t
            \end{bmatrix}
\end{equation}
The allowable twist directions for other possible constraints can be found in \cite{salisbury1985robot}.
\par Given (\ref{eq:J2}) and (\ref{eq:constraint}), the end-effector twist $\Delta{}^0{\mb{x}}_e$ can be expressed as the sum of the twist contribution due to the first kinematic chain moving while the second kinematic chain is locked and the twist contribution of the second kinematic chain while the first kinematic chain is locked. When adding these twists, we take care to transform them into a representation in a frame centered at the end effector point $\mb{e}$ and parallel to $\{0\}$:
\begin{equation}\label{eq:ee_twist_two_chains}
{}^0\Delta{\mb{x}}_e= \mb{S}_{t1} {}^0\Delta\mb{x}_b +  \mb{S}_{t2} {}\Delta^b\mb{x}_{e/b}
\end{equation}
\noindent where the twist transformations $\mb{S}_{t1}$ and $\mb{S}_{t2}$ are given as:
\begin{equation}\label{eq:twists}
\mb{S}_{t1} =
\begin{bmatrix}
\mb{I} & \left[{}^0\mb{b} - {}^0\mb{e}\right]_\times \\
\mb{0} & \mb{I}
\end{bmatrix} \qquad
\mb{S}_{t2} =
\begin{bmatrix}
\Rot{0}{b} & \mb{0} \\
\mb{0} & \Rot{0}{b}
\end{bmatrix}.
\end{equation}
\par Using (\ref{eq:constraint}) and (\ref{eq:J2}), we finally obtain the instantaneous direct kinematics Jacobian for the virtual manipulator comprised of a moving brace-point frame $\{b\}$ and the second kinematic chain:
\begin{equation}\label{eq:braced_kin}
  \Delta{}^0{\mb{x}}_e = \underbrace{\begin{bmatrix}
                     \mb{S}_{t1}\mb{H} & \mb{S}_{t2}\mb{J}_2
                   \end{bmatrix}}_{\mb{A}}\begin{bmatrix}
                                  \dot{\widetilde{\mb{b}}}\\
                                  \dot{\mb{q}}_2
                                \end{bmatrix}
\end{equation}

\section{Compliance Modeling}\label{ch:stiffness}
\par In this section, we present the compliance model used in our simulation results. We assume that the robot is significantly more compliant than the environment such that the compliance of the environment at the bracing location can be neglected, that the kinematic constraints of bracing are not violated, and that contact is always maintained at the bracing point (i.e. no unintentional lift off is allowed).
\subsection{Compliance in Free-Space}
\par To model the compliance of the robot while operating in free space, we use a well-known first-order approximation based on the instantaneous kinematics and statics equations \cite{salisbury_active_1980,huang_duality_2002}. A number of related works consider a similar model that includes second-order effects due to the derivative of the Jacobian \cite{chen_conservative_2000,dumas2011joint,alamdari_stiffness_2018,orekhov_directional_2019}. For cases of sufficiently compliant robots, it has been shown that these second-order effects can be significant \cite{pitt_investigation_2015}, but we leave the study of these second-order effects in the context of bracing for a future work.
\par While in free-space, the compliance of the kinematic chain between frame $\{0\}$ and frame $\{b\}$ is given by:
\begin{equation}\label{eq:C1_def}
\Delta {}^0\mb{x}_b = \mb{C}_1 \Delta {}^0\mb{w}_b, \quad \mb{C}_1 = \mb{J}_1\mb{K}^{-1}_{d_1}\mb{J}_1\T
\end{equation}
where $\Delta{}^0\mb{w}_b = [\Delta{}^0\mb{f}_b\T, \Delta{}^0\mb{m}_b\T]\T \in \realfield{6}$ is a small perturbation in a wrench applied by the robot on the environment at point $\mb{b}$ and $\Delta{}^0\mb{f}_b$, $\Delta{}^0\mb{m}_b$ are the force and moment components of the wrench perturbation from static equilibrium. $\Delta{}^0\mb{x}_b = [\Delta{}^0\mb{p}_b\T, \Delta{}^0\bs{\theta}_b\T]\T \in \realfield{6}$ is a small perturbation in the pose (comprised of a position perturbation $\Delta{}^0\mb{p}_b$ and orientation perturbation $\Delta{}^0\bs{\theta}_b$) of frame $\{b\}$. The matrix $\mb{K}_{d_1} \triangleq \text{diag}(k_{d_1},k_{d_2},\dots,k_{d_n})$ contains the joint-level stiffness values $k_{d_i}$ along its diagonal.
\par Similarly, the compliance of the kinematic chain between frame $\{b\}$ and frame $\{e\}$ is given by:
\begin{equation}\label{eq:C2_def}
\Delta {}^b\mb{x}_{b/e} = \mb{C}_2 \Delta {}^b\mb{w}_e, \quad \mb{C}_2 = \mb{J}_2\mb{K}^{-1}_{d_2}\mb{J}_2\T
\end{equation}
where $\Delta {}^b\mb{x}_{b/e}$ is a small perturbation of frame $\{e\}$ with respect to frame $\{b\}$ and $\Delta{}^b\mb{w}_e = [\Delta{}^b\mb{f}_e\T, \Delta{}^b\mb{m}_e\T]\T \in \realfield{6}$ is a small pertubation of the wrench applied at frame $\{e\}$, expressed in frame $\{b\}$.
\subsection{Bracing-Consistent Compliance of First Kinematic Chain}
Now we consider the compliance of the first kinematic chain while the robot is braced against the environment. A wrench applied to the robot at frame $\{b\}$ must satisfy the statics of the bracing constraint. Under the assumption of frictionless point contact, any change in the applied wrench $\Delta{}^0\mb{w}^*_b$ can be decomposed in directions orthogonal to the constraint and in directions in allowable twist directions:
\begin{equation}
\Delta{}^0\mb{w}^*_b = \underbrace{\mb{P}\Delta{}^0\mb{w}^*_b}_{\text{constraint}} + \underbrace{(\mb{I} - \mb{P})\Delta{}^0\mb{w}^*_b}_{\text{allowable directions}}
\end{equation}
where $\mb{P}$ is a projection matrix along the constraint direction. For a simple single point contact without friction, the reaction wrench and constraint twist are simply a pure force and pure linear velocity. In this case, the projection matrix is trivial and is given by $\mb{P}=\mb{X}\left(\mb{X}\T\mb{X}\right)^{-1}\mb{X}\T$ where \mbox{$\mb{X}=\left[\uvec{n}\T,\mb{0}\T\right]\T\in\realfield{6}$} and $\uvec{n}$ is the local surface normal at the bracing point. In more complex cases involving general constraint wrenches, care must be taken with the proper formulation of this projection as was discussed in \cite{Lipkin1988,Featherstone1999}.
\par The wrench along the constrained direction corresponds to a contact wrench, which is felt by the environment. The change in wrench in the allowable twist direction is a wrench felt by the first kinematic chain:
\begin{equation} \label{eq:wrench_0xb}
\Delta{}^0\mb{w}_b = (\mb{I} - \mb{P})\Delta{}^0\mb{w}^*_b
\end{equation}
If the first portion of the robot were not constrained, frame $\{b\}$ would have experienced a displacement  \mbox{$\Delta{}^0\mb{x}_b = \mb{C}_1\,\Delta{}^0\mb{w}_b$}, as a result of $\Delta{}^0\mb{w}_b$. Because of the bracing constraint, only a portion of $\Delta{}^0\mb{x}_b$ is admissible. We therefore define the admissible deflection consistent with the bracing constraint as $\Delta {}^0\widetilde{\mb{x}}_b$, given by the following:
\begin{equation} \label{eq:twist_0xb_wave_def}
\Delta {}^0\widetilde{\mb{x}}_b = (\mb{I} - \mb{P})\Delta{}^0\mb{x}_b=(\mb{I} - \mb{P})\mb{C}_1\,\Delta{}^0\mb{w}_b
\end{equation}
Defining the \textit{bracing-consistent compliance matrix} as the one relating the kinematically consistent deflection $\Delta {}^0\widetilde{\mb{x}}_b$ with $\Delta{}^0\mb{w}^*_b$ (the total wrench at frame $\{b\}$), i.e.:
 \begin{equation} \label{eq:C1wave_def}
\Delta {}^0\widetilde{\mb{x}}_b = \widetilde{\mb{C}}_1 \Delta {}^0\mb{w}^*_b 
\end{equation}
and substituting (\ref{eq:wrench_0xb}) into (\ref{eq:twist_0xb_wave_def}) results in $\widetilde{\mb{C}}_1$:
\begin{equation} \label{eq:twist_0xb_wave}
\Delta {}^0\widetilde{\mb{x}}_b =\underbrace{(\mb{I} - \mb{P})\mb{C}_1\, (\mb{I} - \mb{P})}_{\widetilde{\mb{C}}_1} \Delta {}^0\mb{w}^*_b
\end{equation}
Therefore, the relationship between the unconstrained compliance $\mb{C}_1$ and $\widetilde{\mb{C}}_1$ is :
\begin{equation} \label{eq:C1wave_detail}
\widetilde{\mb{C}}_1 \triangleq (\mb{I} - \mb{P})\mb{C}_1(\mb{I} - \mb{P})
\end{equation}
\subsection{Bracing-Consistent Compliance of the End-effector}
The instantaneous twist deflections when the robot is subject to a bracing constraint follow the same rationale as in (\ref{eq:ee_twist_two_chains}), except that the twist of frame $\{b\}$ must be kinematically consistent with the bracing constraint:
\begin{equation} \label{eq:tip_disp}
\Delta{}^0\mb{x}_e =\mb{S}_{t1}\Delta{}^0\widetilde{\mb{x}}_b + \mb{S}_{t2}\Delta{}^b\mb{x}_{e/b}
\end{equation}
where $\Delta{}^0\mb{x}_{e/b}$ is a small perturbation of the end-effector with respect to frame $\{b\}$ and $\Delta{}^0\widetilde{\mb{x}}_b$ defined as in (\ref{eq:twist_0xb_wave_def}).
\par Substituting (\ref{eq:C1wave_def}) and (\ref{eq:C2_def}) into (\ref{eq:tip_disp}) gives:
\begin{equation} \label{eq:Ce_before_wrench_sub}
\Delta{}^0\mb{x}_e = \mb{S}_{t1}\widetilde{\mb{C}}_1 \Delta {}^0\mb{w}^*_b + \mb{S}_{t2}\mb{C}_2\Delta {}^b\mb{w}_e
\end{equation}
where $\Delta {}^b\mb{w}_e$ is the end effector wrench expressed in a frame parallel to $\{b\}$ and having its origin at $\mb{e}$ and $\Delta {}^0\mb{w}^*_b$ is the wrench acting on frame $\{b\}$, expressed in a frame parallel to $\{0\}$ and having its origin at $\mb{b}$.
\par The wrenches $\Delta {}^0\mb{w}^*_b$ and $\Delta {}^b\mb{w}_e$ can be related to $\Delta^0\mb{w}_e$ via wrench transformations as the following:  %
\begin{equation}\label{eq:wrench_transoforms}
\Delta{}^0\mb{w}^*_b = \mb{S}_{w1}\,\Delta^0\mb{w}_e, \quad \Delta{}^b\mb{w}_e = \mb{S}_{w2}\Delta^0\mb{w}_e
\end{equation}
where the wrench transformations are given by:
\begin{equation}
\mb{S}_{w1} = \begin{bmatrix}
\mb{I} & \mb{0} \\
\left[{}^0\mb{e} - {}^0\mb{b}\right]_\times & \mb{I}
\end{bmatrix}
\end{equation}
\begin{equation}
\mb{S}_{w2} = \begin{bmatrix}
\Rot{b}{0} & \mb{0} \\
\Rot{b}{0}\left[{}^0\mb{b} - {}^0\mb{e}\right]_\times & \Rot{b}{0}
\end{bmatrix}
\end{equation}
After substituting (\ref{eq:wrench_transoforms}) into (\ref{eq:Ce_before_wrench_sub}) we obtain:
\begin{equation} \label{eq:ee_wrench_after_wrench_sub}
\Delta{}^0\mb{x}_e = \left(\mb{S}_{t1}\widetilde{\mb{C}}_1 \mb{S}_{w1} + \mb{S}_{t2}\mb{C}_2\mb{S}_{w2}\right)\Delta {}^0\mb{w}_e
\end{equation}
Recalling the definition of end-effector compliance as $\Delta{}^0\mb{x}_e = \mb{C}_e\Delta^0\mb{w}_e$, we deduce that the compliance of the end-effector while under a bracing constraint is given by:
\begin{equation}\label{eq:Ce}
\mb{C}_e = \mb{S}_{t1}\widetilde{\mb{C}}_1 \mb{S}_{w1} + \mb{S}_{t2}\mb{C}_2\mb{S}_{w2}
\end{equation}
\subsection{Directional Compliance}
Since specifying a physically realizable stiffness is not trivial, the notion of directional stiffness has been used in \cite{jamshidifar_kinematically-constrained_2017,orekhov_directional_2019}. In the following, we define the directional compliance following the same rationale for directional stiffness.  Using the definition of compliance in (\ref{eq:Ce}), we consider the deflection due to a wrench $\Delta^{0}\mb{w}_e$ with a magnitude $m_w$ acting along a unit screw $\bs{\beta}_w$, i.e.  $\Delta^{0}\mb{w}_e=\bs{\beta}_w m_w$.  Denoting this deflection $\Delta^{0}\mb{x}_e$, we can write:
\begin{equation}\label{eq:delta_x_directional_stiffness}
\Delta{}^0\mb{x}_e = \mb{C}_e \Delta^0{w}_e=\mb{C}_e \bs{\beta}_w m_w
\end{equation}
If the task specification demands a particular directional compliance along a unit twist $\mb{\beta}_x$, the directional compliance $C_{\beta_x}$ is defined such that:
\begin{equation}\label{eq:directional_compliance_basic_def}
\delta_{\beta_x}=C_{\beta_x} m_w
\end{equation}
where $\delta_{\beta_x}$ is the magnitude of deflection along $\mb{\beta}_x$ and it can be expressed in terms of$\Delta{}^0\mb{x}_e$ as:
\begin{equation}\label{eq:delta_beta_as_func_of_delta_x}
\delta_{\beta_x}=\bs{\beta}_x\T \Delta{}^0\mb{x}_e
\end{equation}
Using (\ref{eq:delta_x_directional_stiffness}) in (\ref{eq:delta_beta_as_func_of_delta_x}) and comparing to (\ref{eq:directional_compliance_basic_def}) results in the directional compliance:
\begin{equation}\label{eq:directional_compliance_result}
C_{\beta_x}=\bs{\beta}_x\T \mb{C}_e\bs{\beta}_w
\end{equation}
\par The directional compliance is useful in applications where both the directions of the unit wrench $\bs{\beta}_w$ and the unit twist $\bs{\beta}_x$ are specified. In cases where one is interested in the deflection along $\bs{\beta}_x$ where $\bs{\beta}_w$ could vary or may not be known, a useful stiffness index is:
\begin{equation}\label{eq:compliance_index}
C_i = \| \bs{\beta}_x\T\mb{C}_e  \|
\end{equation}
\section{Redundancy Resolution} \label{ch:red_res}
\par In this section, we describe a redundancy resolution approach for satisfying the kinematic constraints imposed by the bracing point while improving kinematic conditioning and reducing directional compliance. Given a desired end-effector twist $\Delta{}^0\bs{x}_e$, the corresponding configuration velocities $\dot{\widetilde{\mb{b}}}$ and $\dot{\mb{q}}_2$ can be found using the following general solution to (\ref{eq:braced_kin}):

\begin{equation}\label{eq:general_inv}
  \begin{bmatrix}
    \dot{\widetilde{\mb{b}}} \\
    \mb{\dot{q}}_2
  \end{bmatrix} = \left(\mb{A}^+\right)\Delta{}^0\mb{x}_e+\left(\mb{I}-\mb{A}^+\mb{A}\right)\bs{\eta}
\end{equation}

\noindent where $\mb{A}^+$ is the Moore-Penrose pseudoinverse of $\mb{A}$ (which was defined in (\ref{eq:braced_kin})) and $\mb{I}-\mb{A}^+\mb{A}$ projects the vector $\bs{\eta}$ into the null space of $\mb{A}$. Once $\dot{\widetilde{\mb{b}}}$ is found using (\ref{eq:grad_proj}), the corresponding value of $\dot{\mb{q}}_1$ is found using:

\begin{equation}\label{eq:q1}
  \dot{\mb{q}}_1 = \left(\mb{J}_1^{-1}\right)\mb{H}\dot{\widetilde{\mb{b}}}
\end{equation}
\par An appropriate selection of $\bs{\eta}$ in (\ref{eq:general_inv}) allows for the robot to achieve a secondary objective without affecting the desired end-effector motion. Introduced to robotics by Li\'egois in \cite{Liegeois1977}, the gradient projection method locally minimizes an objective function $g$ by selecting $\bs{\eta}=\alpha\nabla g$:

\begin{equation}\label{eq:grad_proj}
  \begin{bmatrix}
    \dot{\widetilde{\mb{b}}}\\
    \mb{\dot{q}}_2
  \end{bmatrix} = \left(\mb{A}^+\right)\Delta{}^0\mb{x}_e+\left(\mb{I}-\mb{A}^+\mb{A}\right)\alpha\nabla g
\end{equation}

\noindent where the scalar $\alpha < 0$ determines the step size for the local optimization which is practically limited by several factors, including joint velocity limits. Methods for selecting an appropriate $\alpha$ can be found in \cite{walker_subtask_1988,Liu2010}.
%
\begin{figure*}[tbp]
    \centering
    \includegraphics[width=0.96\textwidth]{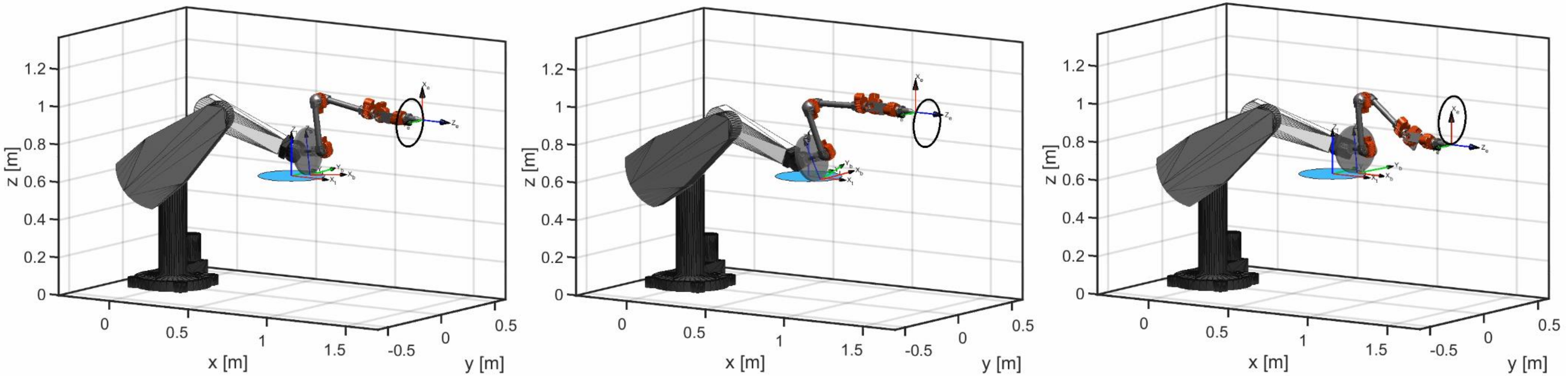}
    \caption{Film strip showing the gradient projection redundancy resolution simulation shown in the multimedia extension.}
    \label{fig:film}
\end{figure*}
\subsection{Defining the Objective Function}
We define the objective function as the sum of weighted performance measures:
\begin{equation}\label{eq:objective}
  g = \alpha_1k+\alpha_2C_i+\alpha_3\theta_z+\alpha_4d
\end{equation}
\noindent where $\alpha_1$, $\alpha_2$, $\alpha_3$, and $\alpha_4$ are positive scalar weights and the measures $k$, $\theta_z$ and $d$ are defined below.
\par The first term of the objective function will attempt to maximize the robot's kinematic isotropy using the Frobenius norm condition number $k$. For isotropic configurations, $k$ is unity and increases to infinity as the robot approaches a singular configuration \cite{khan_kinetostatic_2006}. The Frobenius norm condition number is defined as:
\begin{equation}\label{eq:manip}
  k = \sqrt{\frac{1}{36}\operatorname{Tr}\left(\mb{J}\mb{J}\T\right)\operatorname{Tr}\left(\left(\mb{J}\mb{J}\T\right)^{-1}\right)}
\end{equation}
In this equation, $\operatorname{Tr}\left(\cdot\right)$ is the trace operator and $\mb{J}$ is the free-space Jacobian matrix:
\begin{equation}\label{eq:J}
  \mb{J} = \begin{bmatrix}
             \mb{S}_{t1}\mb{J}_1 & \mb{S}_{t2}\mb{J}_2
           \end{bmatrix}
\end{equation}
The rationale for using the free-space Jacobian is that the entire manipulator should avoid kinematic ill-conditioning in order to effectively be able to satisfy the instantaneous kinematics consistent with bracing constraints.
\noindent Other kinematic conditioning numbers could also be used, a comprehensive survey of which can be found in \cite{Patel2014}.
\par The second term minimizes the directional compliance performance measure $C_i$ from (\ref{eq:compliance_index}). As mentioned above, for some tasks other stiffness performance measures may also be suitable (see \cite{abdolshah_optimizing_2017,anson_orientation_2017,alamdari_stiffness_2018}).
\par The third term of the objective function aims to minimize the angle between the bracing plane normal ${}^0\hat{\mb{z}}_t$ and the vertical axis of the bracing frame ${}^0\hat{\mb{z}}_b$:

\begin{equation}\label{eq:theta}
  \theta_z = \operatorname{acos}\left({}^0\hat{\mb{z}}_t\T{}^0\hat{\mb{z}}_b\right)
\end{equation}

\noindent This term prevents ${}^0\hat{\mb{z}}_b$ from deviating too far from the surface normal, which helps prevent contact with the bracing plane by portions of the robot adjacent to frame $\{b\}$.
\par The last term of the objective function helps prevent frame $\{b\}$ from moving outside the bracing region. The allowable bracing region would be defined in our target application by the geometry of the environment. Here, we assume the allowable bracing region can be modeled as a circle and define a function that penalizes movement away from the center of the circle:
\begin{equation}
d = \frac{r_{max}^2}{r_{max}^2-r^2}
\end{equation}
where $r = \| {}^0\mb{p}_b - {}^0\mb{p}_t\|$ is the distance of frame $\{b\}$ from the constraint frame $\{t\}$ and $r_{max}$ is the maximum allowable radius from the constraint frame. This function grows to infinity as $r$ approaches $r_{max}$. A similar function has historically been used for joint-limit avoidance \cite{chan_weighted_1995}.

\subsection{Calculating $\nabla g$}
The objective function $g$ is a function of the joint variables $\mb{q}_1$ and $\mb{q}_2$. The joint speeds $\dot{\mb{q}}_1$ are a function of the bracing point velocity $\dot{\widetilde{\mb{b}}}$, therefore, for a given initial configuration $\mb{q}_1=\mb{q}_1(\widetilde{\mb{b}})$, $g=g(\mb{q}_1(\widetilde{\mb{b}})),\mb{q}_2)$ and the gradient of $g$ with respect to $\left[\widetilde{\mb{b}}\T, \mb{q}_2\T\right]\T$ is given by:
\begin{equation}\label{eq:g_gradient}
\nabla g=\left[\left(\frac{\partial g}{\partial \mb{q}_1}\frac{d\mb{q}_1}{d\widetilde{\mb{b}}}\right),
\left(\frac{\partial g}{\partial \mb{q}_2}\right)\right]\T
\end{equation}
\noindent where we use the convention that $\frac{\partial g}{\partial \mb{q}_1}$ is a row vector. Using the instantaneous kinematics of the first kinematic chain under bracing constraints:
\begin{equation}\label{eq:q1_h}
  \mb{J}_1\delta\mb{q}_1 = \mb{H}\delta\widetilde{\mb{b}}
\end{equation}
\noindent we obtain:
\begin{equation}\label{eq:dq_dh2}
  \frac{\partial g}{\partial\widetilde{\mb{b}}}= \left(\frac{\partial g}{\partial \mb{q}_1}\right)\mb{J}_1^{-1}\mb{H}
\end{equation}
The gradient of $g$ can now be written as:

\begin{equation}\label{eq:grad_g}
\nabla g=\left[\left(\frac{\partial g}{\partial \mb{q}_1}\right)\mb{J}_1^{-1}\mb{H},
\left(\frac{\partial g}{\partial \mb{q}_2}\right)\right]\T
\end{equation}
In the simulations shown below, $\frac{\partial g}{\partial \mb{q}_1}$ and $\frac{\partial g}{\partial \mb{q}_2}$ are found using central finite differences.
%
\section{Simulation Results}\label{ch:results}
\begin{figure*}[htbp]
    \centering
    \includegraphics[width=0.7\textwidth]{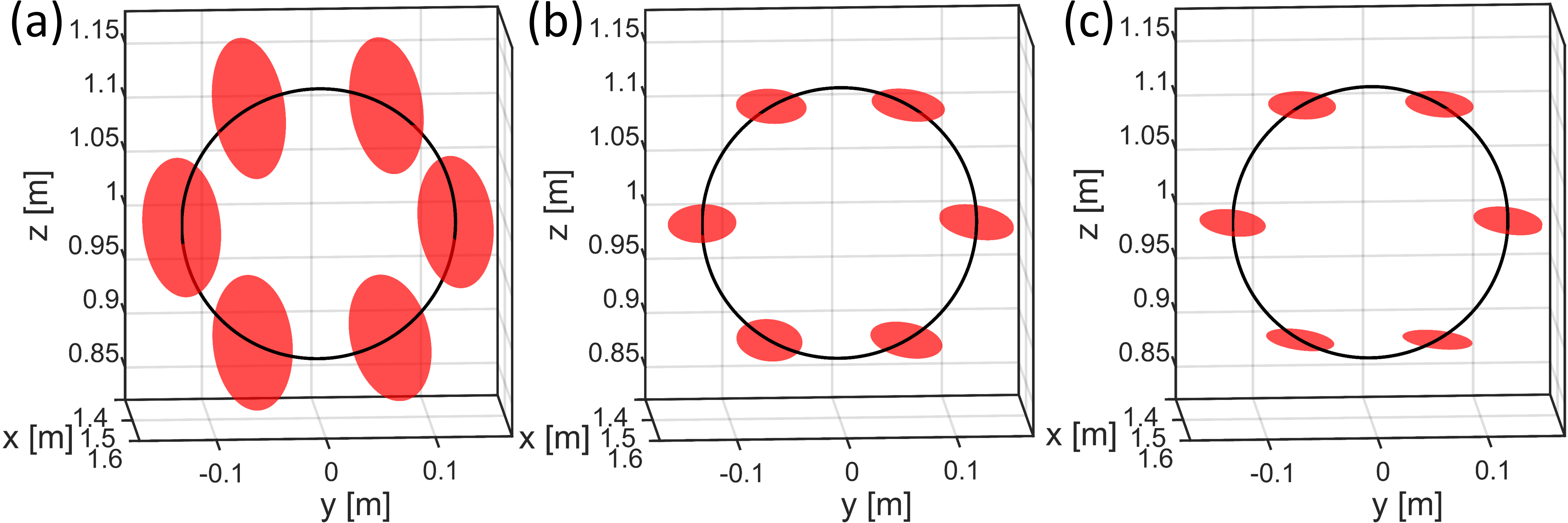}
    \caption{Compliance ellipsoids at representative points along the task for (a) the free space minimum norm, (b) braced minimum norm, and (c) braced gradient projection redundancy resolution simulations.}
    \label{fig:ellip}
\end{figure*}
\begin{figure*}[htbp]
    \centering
    \includegraphics[width=1.0\textwidth]{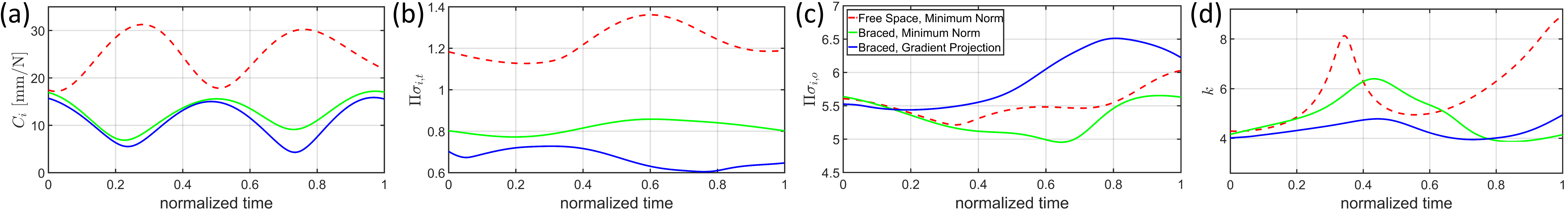}
    \caption{Comparison of (a) $C_i$, (b) $\Pi\sigma_{i,t}$, (c) $\Pi\sigma_{i,o}$, and (d) $k$ for the free space minimum norm, braced minimum norm, and braced gradient projection redundancy resolution methods.}\label{fig:measure}
\end{figure*}
Here we present the results of a kinematic simulation in MATLAB testing the redundancy resolution approach described above. The robot used in the simulation consists of five HEBI Robotics\texttrademark\space X8-16 series elastic actuators attached to the end-effector of a PUMA 560. The elastic element in the HEBI actuators have a stiffness of 170 Nm/rad \cite{hebi}. For the purpose of simulation, the joint stiffness of the PUMA 560 are assumed to also be 170 Nm/rad (which can be assumed to be attainable through a joint-level compliance controller since the PUMA joints are all backdrivable). This robot is used to simulate following a circular path using three different redundancy resolution strategies: 1) free-space minimum norm, 2) minimum norm while maintaining a point bracing contact, and 3) the gradient projection approach described above. While utilizing the gradient projection method, the robot tries to minimize compliance radial to the circle. The other parameters of the gradient projection method are $\alpha = -1$, $\alpha_1 = 0.5$, $\alpha_2 = 100$, $\alpha_3 = 0.6$, and $\alpha_4 = 0.01$. $\nabla g$ was calculated using a finite difference of $\delta q_i = 0.00001$ rad. These gains were selected using a trial-and-error approach.
\par Figure \ref{fig:film} shows a film strip of the robot during the gradient projection simulation shown in the multimedia extension. In this figure, the blue horizontal circle is the bracing surface and the black vertical circle is the task. Figure \ref{fig:measure}(a), shows a comparison of the compliance performance measure $C_i$ for the three redundancy resolution strategies and Fig. \ref{fig:ellip} shows the compliance ellipsoids at representative points along the task. Figures \ref{fig:measure}(b) and \ref{fig:measure}(c) show the product of the singular values of the translational and rotational portion of the braced Jacobian $\widetilde{\mb{J}}$ throughout the task ($\Pi\sigma_{i,t}$ and $\Pi\sigma_{i,o}$, respectively). Lastly, Fig. \ref{fig:measure}(d) shows the Frobenius norm condition number $k$ of $\widetilde{\mb{J}}$ for the three redundancy resolution strategies. The constrained Jacobian matrix used in Fig. \ref{fig:measure} can be found by plugging (\ref{eq:J1}), (\ref{eq:J2}) and (\ref{eq:twist_0xb_wave_def}) into (\ref{eq:tip_disp}):
\begin{equation}\label{eq:J_wave}
  \Delta^0\bs{x}_e = \underbrace{\begin{bmatrix}\mb{S}_{t1}(\mb{I}-\mb{P})\mb{J}_1 & \mb{S}_{t2}\mb{J}_2\end{bmatrix}}_{\widetilde{\mb{J}}}
  \begin{bmatrix} \dot{\mb{q}}_1 \\ \dot{\mb{q}}_2 \end{bmatrix}
\end{equation}
\par Table \ref{tab:summary} summarizes the simulation results for all three redundancy resolution strategies. The results are displayed in terms of the average directional compliance along the path, $\overline{C}_i$, the minimal end effector directional compliance along the path $C_{i_{min}}$, the average product of singular values of the translational/orientational Jacobians, $\overline{\Pi\sigma}_{i,t}$ and $\overline{\Pi\sigma}_{i,o}$, the minimal singular value of translational/orientational Jacobians along the path, $\sigma^{*}_{t_{min}}$ and  $\sigma^{*}_{o_{min}}$, and the average condition number along the path $\overline{k}$.
\par The results from Fig. \ref{fig:ellip}, Fig. \ref{fig:measure}, and Table \ref{tab:summary} show that bracing significantly reduced $C_i$ as compared to the free space simulation. For example, a reduction in average end-effector directional compliance of 49.7\% and 59.2\% was achieved when comparing the two redundancy resolution strategies using bracing to the free-space option. However, bracing negatively impacted average translational and rotational end-effector dexterity when comparing the free space and braced minimum norm strategies as can be seen from the $\overline{\Pi\sigma}_{i,t}$ and $\overline{\Pi\sigma}_{i,o}$ columns in Table \ref{tab:summary}. Due to its inclusion in (\ref{eq:objective}), the gradient projection objective function was able to increase the average kinematic isotropy (indicated by a decrease in $\overline{k}$) compared to both the free space and braced minimum norm strategies. Similar improvements in alternate performance measures may be achieved by including them in the objective function.
\begin{table}[htbp]
\caption{Comparison of performance measures along the path simulated in the multimedia extension}\label{tab:summary}
\centering
\setlength\tabcolsep{1.5pt}
\begin{tabular}{ |l|c|c|c|c|c|c|c| }
 \hline
    {}                      & $\overline{C}_i$ &$C_{i_{min}}$ &$\overline{\Pi\sigma}_{i,t}$ & $\overline{\Pi\sigma}_{i,o}$ & $\sigma^{*}_{t_{min}}$ & $\sigma^{*}_{o_{min}}$ & $\overline{k}$  \\ [2pt] \hline
 Free space min. norm    & 24.90 & 17.16  & 1.22    & 5.50             & 0.29  & 1.08  & 5.84 \\
 Braced min. norm        & 12.52 & 6.86   & 0.81    & 5.30             & 0.29  & 1.01  & 4.87\\
 Braced grad. projection & 10.16 & 4.30   & 0.66    & 5.88             & 0.29  & 1.15  & 4.33 \\
 \hline
\end{tabular}
\end{table}

\section{Conclusion}\label{ch:conclusion}
\par In this paper, we presented a model of the kinematics and first order end-effector compliance of a serial robot with bracing constraints. We also presented a redundancy resolution strategy that is able to reduce the directional end-effector compliance while executing a task and satisfying the bracing constraints. This redundancy resolution strategy was compared to free space minimum norm and braced minimum norm strategies in simulation.

\par Although our results show the benefits of bracing in terms of decreasing end-effector compliance, there are trade-offs such as workspace reduction and reduction in the robot's available degrees-of-freedom for task completion. We assumed the environment and linkages are very rigid in comparison to the joint level stiffness. For applications where this assumption is not reasonable, second order stiffness effects, which were neglected in this paper, may become significant. For these reasons, a thorough analysis of the task and the robot design is needed when deciding to use this redundancy resolution strategy for an application.

\par In future work, we will investigate incorporating second order stiffness effects, finite environment stiffness, and multiple points of contact into the our model. We also plan to study the effects of bracing contact friction and location of the bracing point along the robot. Lastly, we plan to experimentally validate the stiffness model and demonstrate the redundancy resolution strategy on a physical robot.
\balance
\newpage
\bibliographystyle{IEEEtran}
\bibliography{Main}

\begin{thebibliography}{10}
\providecommand{\url}[1]{#1}
\csname url@samestyle\endcsname
\providecommand{\newblock}{\relax}
\providecommand{\bibinfo}[2]{#2}
\providecommand{\BIBentrySTDinterwordspacing}{\spaceskip=0pt\relax}
\providecommand{\BIBentryALTinterwordstretchfactor}{4}
\providecommand{\BIBentryALTinterwordspacing}{\spaceskip=\fontdimen2\font plus
\BIBentryALTinterwordstretchfactor\fontdimen3\font minus
  \fontdimen4\font\relax}
\providecommand{\BIBforeignlanguage}[2]{{%
\expandafter\ifx\csname l@#1\endcsname\relax
\typeout{** WARNING: IEEEtran.bst: No hyphenation pattern has been}%
\typeout{** loaded for the language `#1'. Using the pattern for}%
\typeout{** the default language instead.}%
\else
\language=\csname l@#1\endcsname
\fi
#2}}
\providecommand{\BIBdecl}{\relax}
\BIBdecl

\bibitem{Lorenzini2019}
M.~Lorenzini, W.~Kim, E.~D. Momi, and A.~Ajoudani, ``{A New Overloading Fatigue
  Model for Ergonomic Risk Assessment with Application to Human-Robot
  Collaboration},'' \emph{International Conference on Robotics and Automation},
  pp. 1962--1968, 2019.

\bibitem{Diken1995}
H.~Diken, ``{Effect of Mass Balancing in the Actuator Torques of a
  Manipulator},'' \emph{Mechanism and Machine Theory}, vol.~30, no.~4, pp.
  495--500, 1995.

\bibitem{Fang2019}
C.~Fang, N.~Kashiri, G.~F. Rigano, A.~Ajoudani, and N.~G. Tsagarakis,
  ``{Exploitation of Environment Support Contacts for Manipulation Effort
  Reduction of a Robot Arm},'' \emph{IEEE International Conference on Robotics
  and Automation}, pp. 9502--9508, 2019.

\bibitem{BookBrcing1984}
W.~Book, V.~Sangveraphunsiri, and S.~Le, ``{The bracing strategy for robot
  operation},'' \emph{Theory and practice of robots and manipulators:
  proceedings of RoManSy}, vol.~84, 1984.

\bibitem{HollisBracingAssembly1992}
R.~Hollis and R.~Hammer, ``{Real and virtual coarse-fine robot bracing
  strategies for precision assembly},'' in \emph{Proceedings 1992 IEEE
  International Conference on Robotics and Automation}.\hskip 1em plus 0.5em
  minus 0.4em\relax IEEE Comput. Soc. Press, 1992, pp. 767--774.

\bibitem{Book_bracing_kin1994}
{Jae Young Lew} and W.~Book, ``{Bracing micro/macro manipulators control},'' in
  \emph{Proceedings of the 1994 IEEE International Conference on Robotics and
  Automation}.\hskip 1em plus 0.5em minus 0.4em\relax IEEE Comput. Soc. Press,
  1994, pp. 2362--2368.

\bibitem{Delson1993}
N.~Delson and H.~West, ``{Bracing to Increase the Natural Frequency of a
  Manipulator: Analysis and Design},'' \emph{International Journal of Robotics
  Research}, pp. 560--571, 1993.

\bibitem{Lee1991bracing_ellipsoids}
S.~Lee and S.~Kim, ``{A Self-Reconfigurable Manipulator System with Dextrous
  Bracing Structure},'' \emph{30th Conference on Decision and Control}, pp.
  1033--1038, 1991.

\bibitem{KimDynamicManipBracing1991}
S.~{Kim} and S.~{Lee}, ``Dynamic coordination of a self-reconfigurable
  manipulator system,'' in \emph{[1991] Proceedings of the 30th IEEE Conference
  on Decision and Control}, Dec 1991, pp. 2404--2409 vol.3.

\bibitem{WestAsadaBracingForceControl1985}
H.~West and H.~Asada, ``{A method for the design of hybrid position/Force
  controllers for manipulators constrained by contact with the environment},''
  \emph{Robotics and Automation. Proceedings. 1985 IEEE International
  Conference on}, vol.~2, pp. 251--259, 1985.

\bibitem{FeatherstoneFrictionlessContactHybridControl1999}
R.~Featherstone, S.~Thiebaut, and O.~Khatib, ``{A general contact model for
  dynamically-decoupled force/motion control},'' in \emph{Proceedings 1999 IEEE
  International Conference on Robotics and Automation (Cat. No.99CH36288C)},
  vol.~4, no. May.\hskip 1em plus 0.5em minus 0.4em\relax IEEE, 1999, pp.
  3281--3286.

\bibitem{ParkMultiContactForceControl2006}
J.~Park, ``{Control Strategies for Robots in Contact},'' Ph.D. dissertation,
  Stanford University, 2006.

\bibitem{Wang2010a}
G.~Wang, F.~Yu, M.~Minami, A.~Yanou, and M.~Deng, ``{Multi-Elbows Bracing
  Dynamical Model of Hyper-redundant Mobile Manipulator},'' in \emph{SICE},
  2010.

\bibitem{Wang2010}
G.~Wang and M.~Minami, ``{Modelling and control of hyper-redundancy mobile
  manipulator bracing multi-elbows for high accuracy/low-energy consumption},''
  \emph{SICE Annual Conference 2010, Proceedings of}, pp. 2371--2376, 2010.

\bibitem{Itoshima2011}
M.~Itoshima, Y.~Toda, T.~Maeba, H.~Kataoka, M.~Minami, and A.~Yanou, ``{Energy
  efficiency rate optimization of bracing robot},'' \emph{SICE Annual
  Conference (SICE), 2011 Proceedings of}, pp. 1294--1299, 2011.

\bibitem{Washino2012}
Y.~{Washino}, M.~{Minami}, H.~{Kataoka}, T.~{Matsuno}, A.~{Yanou},
  M.~{Itoshima}, and Y.~{Kobayashi}, ``Hand-trajectory tracking control with
  bracing utilization of mobile redundant manipulator,'' in \emph{2012
  Proceedings of SICE Annual Conference (SICE)}, 2012, pp. 219--224.

\bibitem{yi_geometric_1993}
B.-J. Yi and R.~A. Freeman, ``\BIBforeignlanguage{en}{Geometric analysis of
  antagonistic stiffness in redundantly actuated parallel mechanisms},''
  \emph{\BIBforeignlanguage{en}{Journal of Robotic Systems}}, vol.~10, no.~5,
  pp. 581--603, Jul. 1993.

\bibitem{simaan_geometric_2003}
N.~Simaan and M.~Shoham, ``\BIBforeignlanguage{en}{Geometric {Interpretation}
  of the {Derivatives} of {Parallel} {Robots}’ {Jacobian} {Matrix} {With}
  {Application} to {Stiffness} {Control}},''
  \emph{\BIBforeignlanguage{en}{Journal of Mechanical Design}}, vol. 125,
  no.~1, p.~33, 2003.

\bibitem{jamshidifar_kinematically-constrained_2017}
H.~Jamshidifar, A.~Khajepour, B.~Fidan, and M.~Rushton,
  ``Kinematically-{Constrained} {Redundant} {Cable}-{Driven} {Parallel}
  {Robots}: {Modeling}, {Redundancy} {Analysis}, and {Stiffness}
  {Optimization},'' \emph{IEEE/ASME Transactions on Mechatronics}, vol.~22,
  no.~2, pp. 921--930, Apr. 2017.

\bibitem{kock_parallel_1998}
S.~Kock and W.~Schumacher, ``A parallel x-y manipulator with actuation
  redundancy for high-speed and active-stiffness applications,'' in
  \emph{Proceedings. 1998 {IEEE} {International} {Conference} on {Robotics} and
  {Automation}}, vol.~3, May 1998, pp. 2295--2300 vol.3.

\bibitem{chakarov_study_2004}
D.~Chakarov, ``Study of the antagonistic stiffness of parallel manipulators
  with actuation redundancy,'' \emph{Mechanism and Machine Theory}, vol.~39,
  no.~6, pp. 583--601, Jun. 2004.

\bibitem{muller_stiffness_2006}
A.~M{\"u}ller, ``Stiffness control of redundantly actuated parallel
  manipulators,'' in \emph{Robotics and {Automation}, 2006. {ICRA} 2006.
  {Proceedings} 2006 {IEEE} {International} {Conference} on}.\hskip 1em plus
  0.5em minus 0.4em\relax IEEE, 2006, pp. 1153--1158.

\bibitem{kim_analysis_1997}
W.-K. Kim, J.-Y. Lee, and B.~J. Yi, ``\BIBforeignlanguage{en}{Analysis for a
  planar 3 degree-of-freedom parallel mechanism with actively adjustable
  stiffness characteristics},'' \emph{\BIBforeignlanguage{en}{KSME
  International Journal}}, vol.~11, no.~4, p. 408, Jul. 1997.

\bibitem{lee_upper-body_2014}
J.~Lee, A.~Ajoudani, E.~M. Hoffman, A.~Rocchi, A.~Settimi, M.~Ferrati,
  A.~Bicchi, N.~G. Tsagarakis, and D.~G. Caldwell, ``Upper-body impedance
  control with variable stiffness for a door opening task,'' in \emph{2014
  {IEEE}-{RAS} {International} {Conference} on {Humanoid} {Robots}}, Nov. 2014,
  pp. 713--719.

\bibitem{lim_design_2013}
W.~B. Lim, S.~H. Yeo, G.~Yang, and I.~M. Chen, ``Design and analysis of a
  cable-driven manipulator with variable stiffness,'' in \emph{2013 {IEEE}
  {International} {Conference} on {Robotics} and {Automation}}, May 2013, pp.
  4519--4524.

\bibitem{simaan_stiffness_2003}
N.~Simaan and M.~Shoham, ``\BIBforeignlanguage{en}{Stiffness {Synthesis} of a
  {Variable} {Geometry} {Six}-{Degrees}-of-{Freedom} {Double} {Planar}
  {Parallel} {Robot}},'' \emph{\BIBforeignlanguage{en}{The International
  Journal of Robotics Research}}, vol.~22, no.~9, pp. 757--775, Sep. 2003.

\bibitem{abdolshah_optimizing_2017}
S.~Abdolshah, D.~Zanotto, G.~Rosati, and S.~K. Agrawal, ``Optimizing
  {Stiffness} and {Dexterity} of {Planar} {Adaptive} {Cable}-{Driven}
  {Parallel} {Robots},'' \emph{Journal of Mechanisms and Robotics}, vol.~9,
  no.~3, p. 031004, 2017.

\bibitem{anson_orientation_2017}
M.~Anson, A.~Alamdari, and V.~Krovi, ``Orientation {Workspace} and {Stiffness}
  {Optimization} of {Cable}-{Driven} {Parallel} {Manipulators} {With} {Base}
  {Mobility},'' \emph{Journal of Mechanisms and Robotics}, vol.~9, no.~3, pp.
  031\,011--031\,011--16, Mar. 2017.

\bibitem{alamdari_stiffness_2018}
A.~Alamdari, R.~Haghighi, and V.~Krovi, ``Stiffness {Modulation} in an
  {Elastic} {Articulated}-{Cable} {Leg}-{Orthosis} {Emulator}: {Theory} and
  {Experiment},'' \emph{IEEE Transactions on Robotics}, 2018.

\bibitem{rice2018passive}
J.~J. Rice and J.~M. Schimmels, ``Passive compliance control of redundant
  serial manipulators,'' \emph{Journal of Mechanisms and Robotics}, vol.~10,
  no.~4, p. 041009, 2018.

\bibitem{orekhov_directional_2019}
A.~L. Orekhov and N.~Simaan, ``Directional {Stiffness} {Modulation} of
  {Parallel} {Robots} {With} {Kinematic} {Redundancy} and {Variable}
  {Stiffness} {Joints},'' \emph{Journal of Mechanisms and Robotics}, vol.~11,
  no.~5, p. 051003, Oct. 2019.

\bibitem{Murray1994}
R.~M. Murray, Z.~Li, and S.~S. Sastry, \emph{{A Mathematical Introduction to
  Robotic Manipulation}}.\hskip 1em plus 0.5em minus 0.4em\relax CRC Press,
  1994.

\bibitem{salisbury1985robot}
J.~K. Salisbury and M.~Mason, \emph{Robot hands and the mechanics of
  manipulation}.\hskip 1em plus 0.5em minus 0.4em\relax MIT press Cambridge,
  MA, 1985.

\bibitem{salisbury_active_1980}
J.~K. Salisbury, ``Active stiffness control of a manipulator in cartesian
  coordinates,'' in \emph{1980 19th {IEEE} {Conference} on {Decision} and
  {Control} including the {Symposium} on {Adaptive} {Processes}}, Dec. 1980,
  pp. 95--100.

\bibitem{huang_duality_2002}
S.~Huang and J.~M. Schimmels, ``\BIBforeignlanguage{en}{The {Duality} in
  {Spatial} {Stiffness} and {Compliance} as {Realized} in {Parallel} and
  {Serial} {Elastic} {Mechanisms}},'' \emph{\BIBforeignlanguage{en}{Journal of
  Dynamic Systems, Measurement, and Control}}, vol. 124, no.~1, p.~76, 2002.

\bibitem{chen_conservative_2000}
S.-F. Chen and I.~Kao, ``\BIBforeignlanguage{en}{Conservative {Congruence}
  {Transformation} for {Joint} and {Cartesian} {Stiffness} {Matrices} of
  {Robotic} {Hands} and {Fingers}},'' \emph{\BIBforeignlanguage{en}{The
  International Journal of Robotics Research}}, vol.~19, no.~9, pp. 835--847,
  Sep. 2000.

\bibitem{dumas2011joint}
C.~Dumas, S.~Caro, S.~Garnier, and B.~Furet, ``Joint stiffness identification
  of six-revolute industrial serial robots,'' \emph{Robotics and
  Computer-Integrated Manufacturing}, vol.~27, no.~4, pp. 881--888, 2011.

\bibitem{pitt_investigation_2015}
E.~B. Pitt, N.~Simaan, and E.~J. Barth, ``An investigation of stiffness
  modulation limits in a pneumatically actuated parallel robot with actuation
  redundancy,'' in \emph{{ASME}/{BATH} 2015 {Symposium} on {Fluid} {Power} and
  {Motion} {Control}}.\hskip 1em plus 0.5em minus 0.4em\relax American Society
  of Mechanical Engineers, 2015.

\bibitem{Lipkin1988}
H.~Lipkin and J.~Duffy, ``{Hybrid Twist and Wrench Control for a Robotic
  Manipulator},'' \emph{Journal of Mechanisms, Transmissions, and Automation in
  Design}, vol. 110, pp. 138--144, 1988.

\bibitem{Featherstone1999}
R.~Featherstone, ``{A General Contact Model for Dynamically-Decoupled
  Force/Motion Control},'' in \emph{International Conference on Robotics {\&}
  Automation}, no. May, 1999, pp. 3281--3286.

\bibitem{Liegeois1977}
A.~Liegeois, ``{Automatic Supervisory Control of the Configuration and Behavior
  of Multibody Mechanisms},'' \emph{IEEE Transactions on Systems, Man, and
  Cybernetics}, vol.~7, no.~12, pp. 868--871, 1977.

\bibitem{walker_subtask_1988}
I.~D. Walker and S.~I. Marcus, ``Subtask performance by redundancy resolution
  for redundant robot manipulators,'' \emph{IEEE Journal on Robotics and
  Automation}, vol.~4, no.~3, pp. 350--354, Jun. 1988.

\bibitem{Liu2010}
Y.~Liu, J.~Zhao, and B.~Xie, ``{Obstacle avoidance for redundant manipulators
  based on a novel gradient projection method with a functional scalar},''
  \emph{2010 IEEE International Conference on Robotics and Biomimetics, ROBIO
  2010}, pp. 1704--1709, 2010.

\bibitem{khan_kinetostatic_2006}
W.~A. Khan and J.~Angeles, ``\BIBforeignlanguage{en}{The {Kinetostatic}
  {Optimization} of {Robotic} {Manipulators}: {The} {Inverse} and the {Direct}
  {Problems}},'' \emph{\BIBforeignlanguage{en}{Journal of Mechanical Design}},
  vol. 128, no.~1, p. 168, 2006.

\bibitem{Patel2014}
S.~Patel and T.~Sobh, ``{Manipulator Performance Measures - A Comprehensive
  Literature Survey},'' \emph{Journal of Intelligent and Robotic Systems:
  Theory and Applications}, vol.~77, no. 3-4, pp. 547--570, 2014.

\bibitem{chan_weighted_1995}
T.~F. Chan and R.~V. Dubey, ``A weighted least-norm solution based scheme for
  avoiding joint limits for redundant joint manipulators,'' \emph{IEEE
  Transactions on Robotics and Automation}, vol.~11, no.~2, pp. 286--292, Apr.
  1995.

\bibitem{hebi}
{HEBI Robotics, Inc}, ``Hebi robotics: Documentation. [online]. available
  \url{http://docs.hebi.us/hardware.html#hw_docs}.''

\end{thebibliography}
\end{document}